# Generative AI in Ship Design

[a] **Sahil Thakur,** [a] **Navneet V Saxena, and** [b,*] **Prof Sitikantha Roy**

[a] *Naval Construction Wing, Department of Applied Mechanics, IIT Delhi, India*
*\*Department of Applied Mechanics, IIT Delhi, India*

The process of ship design is intricate, heavily influenced by the hull form which accounts for approximately 70% of the total cost. Traditional methods rely on human-driven iterative processes based on naval architecture principles and engineering analysis. In contrast, generative AI presents a novel approach, utilizing computational algorithms rooted in machine learning and artificial intelligence to optimize ship hull design. This report outlines the systematic creation of a generative AI for this purpose, involving steps such as dataset collection, model architecture selection, training, and validation. Utilizing the "SHIP-D" dataset, consisting of 30,000 hull forms, the report adopts the Gaussian Mixture Model (GMM) as the generative model architecture. GMMs offer a statistical framework to analyze data distribution, crucial for generating innovative ship designs efficiently. Overall, this approach holds promise in revolutionizing ship design by exploring a broader design space and integrating multidisciplinary optimization objectives effectively.

**Keywords:** Ship design; hull form; generative AI; computational algorithms; Gaussian Mixture Model; dataset optimization; multidisciplinary

## 1. Introduction

Designing a ship is one of the most complex and time-consuming process. From laying down the keel to commissioning, the designing and building of a destroyer approximately takes 7-8 years in India [19]. The most important step in designing a ship is determining the shape of the hull form [3]. Also, the shape of the hull has a direct impact of 70% on the cost of the ship [1]. The conventional approach for designing the hull of a ship involves several key steps like, defining requirements which involves understanding the purpose and operational requirements of the ship, including speed, capacity, range, and environmental conditions. After which the designer develops rough sketches and basic layouts of the hull shape based on requirements and initial feasibility assessments [2].

Then an initial hull form is generated based on the established guidelines and empirical formulas. Hydrodynamic analysis is conducted to evaluate the performance the of the hull design, which also involves assessing parameters such as resistance, seakeeping behavior and stability. Then this hull form is iteratively refined based on the insights gained from this analysis. The conventional form of optimization approach relies on human expertise, naval architecture principles and engineering analysis to modify and optimize the design [1].

In contrast to the conventional optimization approach, the generative approach to design the hull form, involves the use of computational algorithms, which are based on machine learning and AI techniques such as Genetic Algorithms, Neural Networks, and Generative Adversarial Networks (GANs). It is different from the conventional approach as it makes use of large dataset of existing designs, performance data and simulation results, to learn the associated patterns and relationships within the dataset. This feature allows the generation of novel designs. The generative algorithm can explore a broader design space as compared to the conventional methods. It can generate unconventional and innovative designs that a designer might not even think about.

As the generative algorithm is based on machine learning it is able iteratively refine the designs automatically and in a lesser time. This significantly reduces the time and effort to optimize the design. The generative algorithm can also learn from the feedback given to it, thereby improving its performance.

The major advantage of the generative approach over the conventional approach is having a multidisciplinary optimization. Generative algorithms can integrate multiple design objectives and constraints across various disciplines like hydrodynamics, structures, and stability, into a unified optimization network. Thus, generative approach offers a significant advantage in terms of efficiency.

## 2. Overview

Artificial Intelligence (AI) refers to the simulation of human intelligence processes by machines, especially computer systems. It encompasses various approaches and techniques aimed at enabling machines to perform tasks that typically require human intelligence, such as reasoning, problem-solving, learning, perception, understanding natural language, and more [14].

Generative AI is a specific subfield of artificial intelligence focused on creating or generating new content, often in the form of images, text, music, or other types of media. Instead of simply analyzing and processing existing data, generative AI systems are designed to generate novel outputs based on patterns learned from large datasets.

One of the key techniques used in generative AI is Generative Adversarial Networks (GANs), where two neural networks, known as the generator and the discriminator, are trained simultaneously. The generator generates synthetic data while the discriminator evaluates the authenticity of that data. Through this adversarial process, the generator learns to create increasingly realistic outputs.

Generative AI has applications across various domains, including:
- Creative Content Generation: Generative AI can create artwork, music compositions, or even generate new designs for products.
- Data Augmentation: It can be used to generate synthetic data for training machine learning models, thereby expanding the available dataset, and improving model performance.

- Text Generation: Generative models like GPT (Generative Pre-trained Transformer) can generate human-like text, enabling applications such as dialogue systems, content creation, and language translation.
- Image Synthesis and Manipulation: Generative models can generate photorealistic images, alter existing images, or even create entirely new scenes.
- Drug Discovery and Molecular Design: Generative models can assist in generating new molecules with desired properties for drug discovery and material science.

### 3. Steps in creating Gen AI

After clearly defining the objectives and scope of the generative AI, the following steps are followed to create a Gen AI [13]:

- Data Collection: To train the generative model, relevant datasets are gathered. These datasets should be sufficiently diverse and large to capture the variability in the data.
- Data Preprocessing: The collected data is cleaned and preprocessed to remove duplicate entries, handle missing values, standardize format, and ensure consistency.
- Choosing a Generative Model Architecture: The architecture defines the structure of the model. This means that it impacts the way the AI model analyses the training data and how it uses this data to create new content.
- Model Training: The selected generative model is trained using pre-processed data. This also involves optimizing the model parameters to minimize the loss function. The loss function measures the difference between the generated outputs and the realistic data.
- Model Optimization: This is also known as Hyperparameter Tuning. Hyperparameter tuning is like adjusting the settings of a Machine Learning model to make it work better. These settings are called hyperparameters which are like dials controlling how the model learns from the dataset. By fine-tuning hyperparameters such as the learning rate, batch size etc., the model's performance can be optimized and the rate at which it learns from the dataset.
- Evaluation: Evaluate and validate the generated output by comparing it with realistic data and domain knowledge.

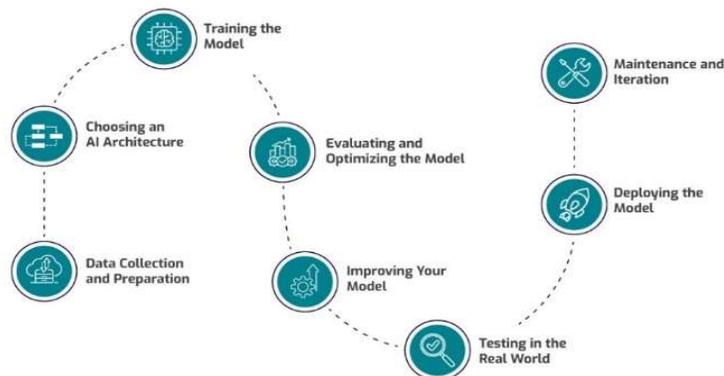

*Figure 1: Flow of building a Gen AI model*

## 4. Dataset

For understanding and generating a novel hull form design using generative AI, the dataset used for training the generative model was taken from the work published by Noah J. Bagazinski and Faez Ahmed, in their paper "SHIP-D: Ship Hull Dataset for Design Optimization Using Machine Learning." [1] Thirty thousand hull forms with design and functional performance data, including mesh, parameterization, and point-cloud, make up the content of the dataset [1]. Each hull form is defined by 45 parameters. This dataset was chosen as it consists of design representation which is comprehensive enough to cover all the features of traditional hull forms. Previously, dimensionality analysis performed on different parent hull forms, undertaken by Wang et al. and Kahn et al, gave the results that 32 and 27 parameters can construct a reasonably complex surface feature of hulls. For getting a diverse spread of hull forms, the dataset chosen consist of hull forms defined by 45 parameters.

The 45 parameters are as defined below:

- Principal dimensions of the hull are defined by seven terms which are length, beam, depth, draft, and tapers at the ends of the hull.

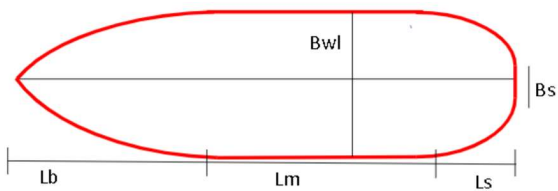

*Figure 2: Principal Dimensions*

- The cross section of the hull in the parallel midbody is defined by four terms. These parameters can produce cross sections of the flare, tumblehome, bilges, chines, and S-chines that are visible on classic hulls.

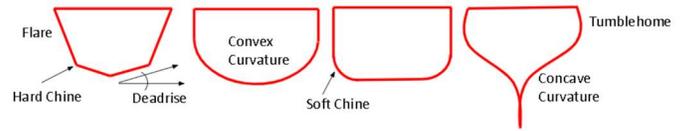

*Figure 3: Cross section of the hull*

- The tapered regions at the bow and stern of the hull are defined by a total of twenty terms. These parametric terms define features such as drift angle, keel rise, transom cross section, rake, and the transition from the taper to the parallel middle body.

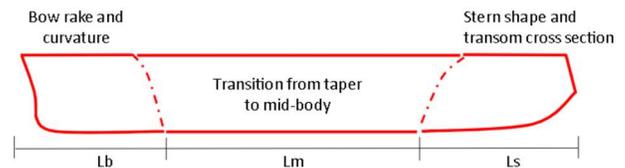

*Figure 4: Tapered Regions*

- The bow and stern bulb geometries are defined by fourteen terms, including terms that define the size, vertical asymmetry, and the fillet transition of the bulb into the hull.

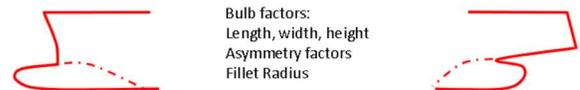

*Figure 5: Bow and Stern bulb geometries*

The hull forms defined by the parameters in the dataset were randomly generated and consists of three distinct subsets of hulls [1]. If the randomly generated parameters resulted in a feasible hull shape, then only it was included in the dataset. This way of generating hull forms was repeated until each subset comprised ten thousand hulls, making a total of thirty thousand hulls in the dataset.

The first part of the dataset comprises hulls created with a broad range of parameters, resulting in diverse combinations of geometric features. These hulls represent various possible designs. In the second part, hulls without bulbs are emphasized, showcasing characteristics commonly found in smaller vessels like tugboats, fishing trawlers, and yachts. The third part focuses on hulls biased towards features typical of larger vessels, such as warships and cargo ships. These hulls have specific parameters like a positive keel radius and a deadrise angle of zero degrees.

Defining and organizing the dataset in this manner ensures representation of features similar to real-world hulls that are of different sizes and can be classified under various types of vessels.

The parameters that define each hull form is shown in the table below:

*Table 1: Parameters*

| Hull Section | Variable | Name | Units/Scaling Measure | Value Range in Dataset |
|---|---|---|---|---|
| **Principle Dimensions** | LOA | Length Overall | Primary measure of ship's scale in meters | LOA = 10 |
| | Lb | Length of Bow Taper | Fraction of LOA | $0.05 < Lb < 0.9$ |
| | Ls | Length of Stern Taper | Fraction of LOA | $0.0 < Ls < 0.9$ |
| | Bd | Beam at Midship Deck | Fraction of LOA | $0.0833 < Bd < 0.333$ |
| | Dd | Depth of Hull | Fraction of LOA | $0.05 < Dd < 0.25$ |
| | Bs | Beam at Stern Deck | Fraction of Bd | $0.0 < Bs < 1.0$ |
| | WL | Design Draft | Fraction of Dd, used for bulb design | $0.0 < WL < 1$ |
| **Midship Cross Section** | Bc | Beam at Chine | Fraction of LOA | $0.05 < Bc < 0.5$ |
| | Beta | Deadrise angle | Degrees | $0.0 < Beta < 45.0$ |
| | Rc | Radius of Chine | Fraction of Bc (strictly positive) | $0.0 < Rc < 1.0$ |
| | Rk | Radius of Keel | Fraction of Dd (can be positive or negative) | $-1.0 < Rk < 1.0$ |
| **Bow Geometry** | BOW(A) | Constants for Parabolic Bow Shape | $BOW(z) = Az^2 + Bz + C$, where C is solved so that $min(Bow(z)) = 0$, A, B, and C are | $-4.0 < BOW(A) < 4.0$ |
| | BOW(B) | | | $-4.0 < BOW(B) < 4.0$ |
| | | | scaled by Lbb and Dd | |
| | BK | Bow-Keel Intersect | Fraction of Dd where Bow curve and keelrise curve intersect | 0.0 < BK < 1.0 |
| | Kappa_BOW | Start of keelrise – Bow | Fraction of Lb, where keel-rises from Z =0 towards bow | 0.0 < Kappa_Bow < 1.0 |
| | DELTA_BOW(A) | Constants to define curve for midship width - | DELTA_BOW(z) = $Az^2+Bz+C$, where C is solved algebraically. Defines x position where midship beam for given z is achieved along bow taper | -4.0 < DELTA_BOW(A) < 4.0 |
| | DELTA_BOW(B) | | | 4.0 < DELTA_BOW(B) < 4.0 |
| | DRIFT(A) | Constants for curve that define drift angle along BOW(z) | DRIFT(z) = $Az^2+Bz+C$, defines the drift angle in degrees from the bow as a function of height. | -4.0 < DRIFT(A) < 4.0 |
| | DRIFT(B) | | | -4.0 < DRIFT(B) < 4.0 |
| | DRIFT(C) | | | 0 < DRIFT(C) < 60 |
| **Stern Geometry** | bit_EP_S | Lower stern taper bit | Defines if stern taper is (1) Ellipse or (0) Parabola below transom | 1 or 0 |
| | bit_EP_T | Upper stern taper bit | Defines if stern taper is (1) Ellipse of (0) Parabola for the transom | 1 or 0 |
| | TRANS(A) | Transom Slope | Transom(z) = $Az + B$, defines the transom position between Dd and SK | -3.0 < TRANS(A) < 5.0 |
| | SK | Stern-Keel Intersect | Defines intersection of Transom and the keelrise for the stern, fraction of Dd | 0.0 < SK < 1.0 |
| | Kappa_STERN | Start of keelrise – stern | Fraction of Ls where keel rises from z = 0 towards transom | 0.0 < Kappa_STERN < 1.0 |
| | DELTA_STERN(A) | Constants to define curve for midship width - | DELTA_STERN(z) = Az2+Bz+C, where C is solved algebraically. Defines x position where midship beam for given z is achieved along stern taper | -4.0 < DELTA_STERN(A) < 4.0 |
| | DELTA_STERN(B) | | | 4.0 < DELTA_STERN(B) < 4.0 |
| | Beta_trans | Deadrise angle for transom | Degrees | 0 < Beta_trans < 60 |
| | Bc_trans | Beam at Transom Chine | Fraction of LOA | 0 < Bc_trans < 0.5 |
| | Rc_trans | Transom Chine Radius | Fraction of Bc_trans | 0 < Rc. Trans < 0.5 |
| | Rk_trans | Transom Keel Radius | Fraction of Dd*(1-SK) | -1.0 < Rk_trans < 1.0 |
| **Bulb Geometries** | bit_BB | Bulbous Bow Bit | Defines if (1) there is a bulbous bow or (0) not | 1 or 0 |
| | bit_SB | Bulbous Stern Bit | Defines if (1) there is a bulbous bow or (0) not | 1 or 0 |
| | Lbb | Length of Bulbous Bow | Fraction of LOA | 0.0 < Lbb < 0.2 |
| | Hbb | Height of BB Max Length | Fraction of WL*Dd | 0.0 < Hbb < 1.0 |
| | Bbb | Beam of BB | Fraction of Beam at z = Hbb | 0.0 < Bbb < 1.0 |
| | Lbbm | Length of Long. Bulb Curvature | Fraction of Lbb where Bulb curve begins | -1.0 < Lbbm < 1.0 |
| | Rbb | Fillet Radius for BB | Defines fillet radius of BB-Bow intersect as a fraction of Lbb | 0.05 < Rbb < 0.33 |
| | Kappa_SB | Start Position of Stern Bulb | Defines x position of Stern Bulb as a fraction of Lb | 0.0 < Kappa_SB < 1.0 |
| | Lsb | Length of Stern Bulb | Fraction of LOA | 0.0 < Lsb < 0.2 |
| | HsbOA | Height overall of Stern Bulb | Fraction of WL*Dd | 0.0 < HsbOA < 1.0 |
| | Hsb | Height of SB Max Length | Fraction of HsbOA*WL*Dd | 0.0 < Hsb < 1.0 |
| | Bsb | Beam of SB | Fraction of Beam at z = Hsb | 0.0 < Bsb < 1.0 |

| | Lsbm | Length of Long. Bulb Curvature | Fraction of Lsb where Bulb curve begins | -1.0 < Lsbm < 1.0 |
|---|---|---|---|---|
| | Rsb | Fillet Radius for SB | Defines fillet radius of SB-Stern Intersect as a fraction of Lsb | 0.05 < Rsb < 0.33 |

The aim of generating the parameterized hulls randomly was to construct a dataset that comprehensively covers the potential design variations of ship hulls meeting feasibility criteria. This approach allows a machine learning model to understand the relative effectiveness of individual and combined geometric features of a hull.

## 5. Clustering

K-means clustering is a popular method for clustering unlabeled datasets. K-means is an iterative algorithm that divides a dataset into K number of distinct and non-overlapping clusters. First, it is necessary to specify the number of clusters that will be created [16].

The elbow method and silhouette method are two commonly used techniques for determining the optimal number of clusters in an unlabeled dataset. The elbow method involves plotting the within-cluster sum of squares (WCSS) against the number of clusters and identifying the "elbow" point where the rate of decrease in WCSS slows down significantly. This point can be taken as the optimal number of clusters that are present in the dataset, as adding more clusters beyond this point will not significantly reduce the WCSS distance.

The silhouette method measures how similar a data point is to its own cluster when compared to other clusters. It computes the silhouette coefficient for each data point and averages it across all data points for different numbers of clusters. A higher silhouette score suggests better-defined clusters. By comparing silhouette scores across different numbers of clusters, the optimal number of clusters can be identified where the silhouette score is highest. Both methods provide valuable insights into the appropriate number of clusters, allowing for more informed decision-making in clustering analysis.

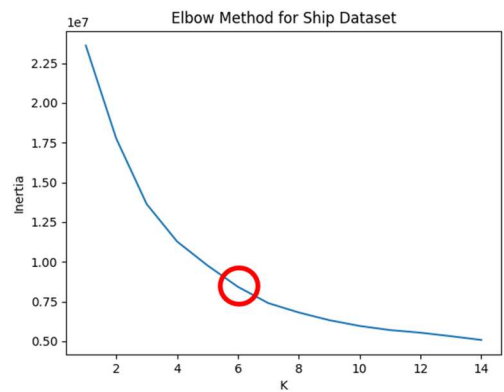

*Figure 7: Elbow Method*

Then, randomly initialize K cluster centroids (points representing the center of each cluster) in the feature space. Each data point is then assigned to the nearest centroid based on some distance metric, typically Euclidean distance. This step creates clusters where data points are closest to the centroid of their respective cluster.

After all data points have been assigned to clusters, the centroids are updated by calculating the mean of all data points assigned to each cluster. The centroids are recalculated based on the current assignment of data points. The above steps are repeated iteratively until convergence is reached. It can be said that convergence occurs when the centroids are no longer changing significantly between iterations or else when a specified number of iterations is reached. The result is a set of K clusters, where each data point belongs to the cluster whose centroid it is closest to.

K-means is computationally efficient and works well with large datasets, making it widely used in various applications such as customer segmentation, image compression, and document clustering. However, it requires the number of clusters, K, to be specified beforehand, and it assumes clusters are spherical and of similar size, which may not always be the case in real-world data.

A python code was written to apply K-means algorithm to the dataset. The result of which indicated 06 optimal number of clusters in the dataset.

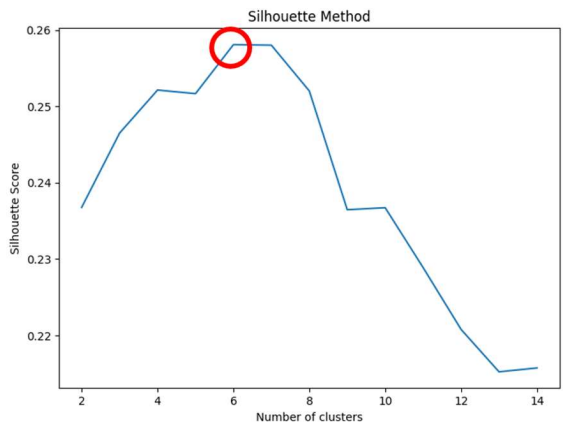

*Figure 8: Silhouette Method*

## t-SNE

It is a dimensionality reduction technique commonly used for visualizing high-dimensional data, particularly in scenarios where datasets are represented by a large number of parameters [17]. Unlike traditional techniques such as PCA (Principal Component Analysis), t-SNE preserves local similarities in the data, making it well-suited for capturing complex structures and patterns in high-dimensional datasets.

By transforming high-dimensional data into a lower-dimensional space while preserving local relationships, t-SNE enables the visualization of large datasets in two or three dimensions, making it easier to interpret and analyze.

This makes t-SNE particularly useful for exploring and understanding complex datasets with a large number of parameters, such as those encountered in fields like genomics, image processing, and natural language processing. However, it is important to note that t-SNE is computationally intensive and may not always preserve global structures, so interpretation should be done cautiously, especially with large datasets [17].

t-SNE algorithm was applied on the dataset using a python code, and a plot was generated by reducing the 45 parameters which defines each hull form, to a dimension of two features only.

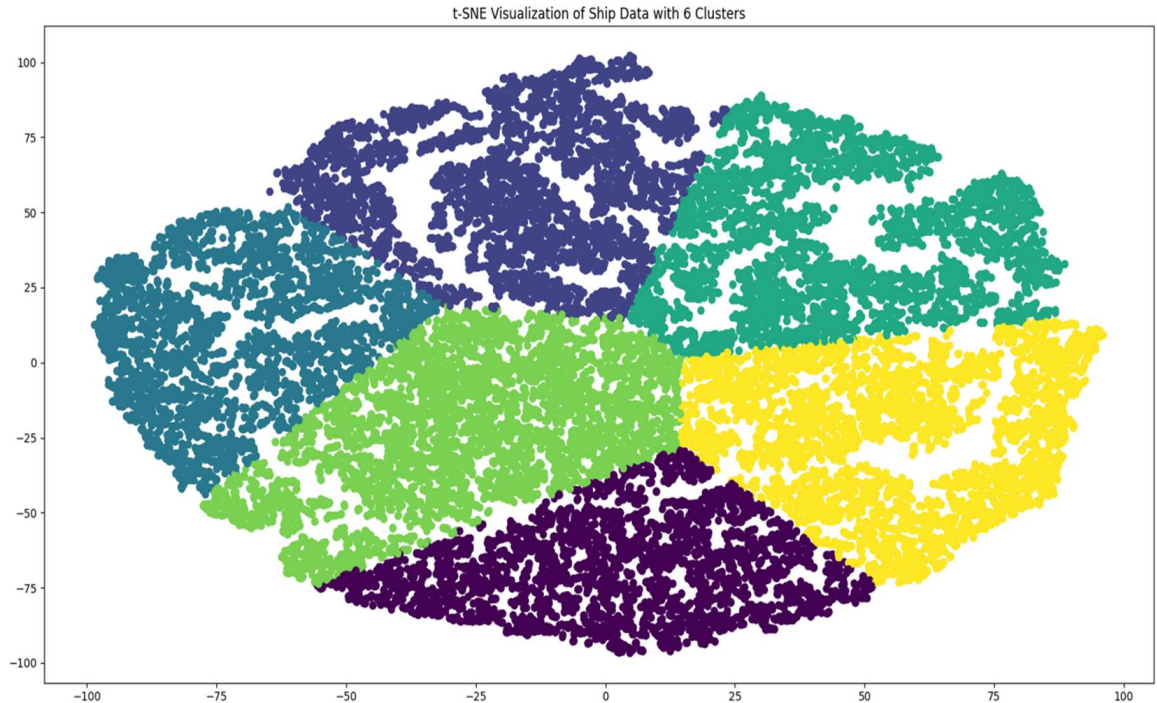

*Figure 9: Visualization of the dataset*

## 6. Model Architecture

A generative model is designed to understand the hidden and associated patterns in the dataset and reproduce the features or characteristics of the data. An example of this would be training the generative model with pictures of cats. The model's task is not just to identify that whether the image projected is that of a cat or not, but rather to learn the features that makes up that image of cat. Example of features in this case would be their shape, colours, patterns and textures. After understanding the features, it is able to generate new images of cats that looks realistic but is not a replica of any images that the generative model has been trained on from the training dataset.

Generative architecture stands as a pioneering force within contemporary design, leveraging cutting-edge technologies to revolutionize the process of design conception and implementation. Within its arsenal of methodologies lie diverse tools such as Cellular Automata, which adeptly generate intricate designs through fundamental rules, and Shape Grammar, a framework governed by stringent principles to delineate architectural forms [4]. Additionally, Genetic Algorithms draw inspiration from natural processes to ascertain optimal solutions for intricate design challenges, while Artificial Neural Networks emulate cognitive functions, seamlessly navigating through vast datasets [5]. Furthermore, methodologies like Finite Element Analysis and Topology Optimization contribute to the enhancement of structural integrity and efficiency. Collectively, these tools facilitate the exploration of conceptual realms, automate routine tasks, and amplify creative endeavors within the realm of architecture [6]. The generative model architecture employed in this study is the Gaussian Mixture Model (GMM).

A Gaussian Mixture Model (GMM) is a mathematical tool used to sort data into different groups, based on their way of distribution. It is like organizing different types of items into separate bins. This model is handy because it can figure out these groups without needing to know anything about them beforehand. GMMs are used in lots of areas, like analyzing pictures, tracking objects, and pulling important features out of speech. The model works on the idea that the data comes from a mix of different patterns [7].

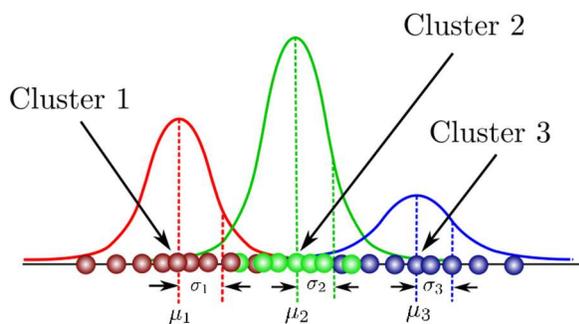

*Figure 10: Gaussian distributions*

The Gaussian Mixture Model (GMM) serves as a statistical framework to explain the distribution of data, by creating a fusion of numerous Gaussian distributions, each defined by distinct mean and variance parameters. Its application holds particular significance in machine learning contexts where intricate and multi-dimensional datasets are prevalent [7]. The key concepts associated with GMMs are:

- Gaussian Distribution: GMM assumes that the data points within each cluster are generated from a Gaussian (normal) distribution. A Gaussian distribution is characterized by its mean ($\mu$) and covariance ($\Sigma$).

- Mixture Model: GMM represents the overall distribution of the data as a mixture of multiple Gaussian distributions, each associated with a different cluster. The model calculates the probability of a data point belonging to each cluster.

- Cluster Parameters: For each cluster in the mixture, GMM estimates parameters including the mean ($\mu$) and covariance ($\Sigma$) of the Gaussian distribution. These parameters are optimized during the training process to best fit the observed data.

- Cluster Assignment: GMM assigns each data point to one of the clusters probabilistically based on the likelihood of the data point belonging to each cluster. This is computed using Bayes' theorem.

- Expectation-Maximization (EM) Algorithm: GMMs are typically trained using the EM algorithm. In the E-step, the algorithm estimates the probabilities of cluster assignments given the data and the current parameters. In the M-step, it updates the parameters (means and covariances) based on these assignments.

- Initialization: GMM training often requires an initial guess for the parameters. Common initialization methods include random initialization and K-means clustering.

- Number of Clusters (K): The number of clusters in a GMM needs to be specified in advance. Determining the optimal number of clusters often involves techniques such as the elbow method or cross-validation.

- Likelihood Maximization: GMM training aims to maximize the likelihood of observing the data given the model parameters. This is typically achieved through iterative optimization techniques like gradient ascent.

- Soft Clustering: Unlike hard clustering algorithms (such as K-means), which assign each data point to a single cluster, GMM provides soft assignments where data points can belong to multiple clusters with varying probabilities.

### 7. Generating Novel Hull Design

A novel design with regards to Gaussian Mixture Models (GMM) typically refers to the creation of new data points that closely resemble the distribution of the original dataset but are not exact replicas of any existing data points.

Generating novel designs involves sampling from the learned GMM to produce new data points that exhibit similar characteristics to the original dataset but introduce variations or combinations not explicitly present in the training data. A python code was written so that the GMM is trained on this pre-processed dataset, learning the underlying patterns and structures inherent in the data. Post-training, the model can generate new designs by sampling from the learned probability distribution, resulting in creations that exhibit similarities to the training data but feature novel combinations and variations. To ensure that the model generated is novel, another code was written to check its uniqueness by comparing it with the hull forms on which the generative model was trained.

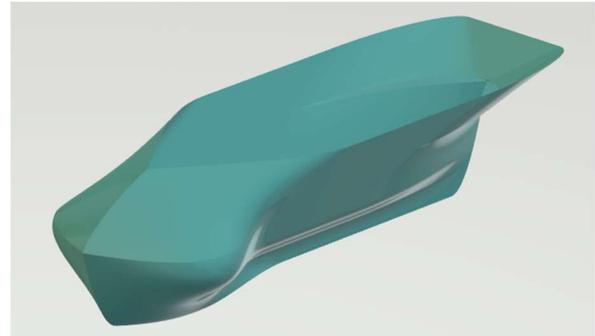

Figure 12: Novel Design

Another important aspect to new generated hull form is to check whether it is a feasible design or not. To meet the requirements of a "feasible" hull shape, the surface of the hull must adhere to two conditions:

- The hull must be watertight, ensuring there are no openings or gaps in its surface.
- The hull surface must avoid self-intersection, meaning it does not intersect with itself.

As the hull surface is described by equations incorporating constants influenced by parameter values, assessing whether a hull surface meets the primary feasibility criteria can be achieved through algebraic solutions. The benefit of solving these conditions algebraically lies in the substantial reduction of computational resources required compared to feasibility checks involving mesh generation [1]. Since the dataset already consists of hull forms on which this feasibility check is already done, there is no need to check the feasibility criteria on the new generated hull form.

Extracting novel designs after training a Gaussian Mixture Model (GMM) and determining the optimal number of components can be achieved through several methods:

- Sampling: Randomly drawing samples from the learned GMM according to their respective weights. This process generates new designs based on the distribution learned from the dataset.
- Prototype Generation: Identifying the means of Gaussian components representing unique or interesting designs. These means act as prototypes for different design clusters, allowing for the extraction of novel designs through modification or combination of these prototypes.
- Interpolation: Employing interpolation techniques within the latent space of the GMM to generate designs that lie between existing designs. This involves traversing along the axes defined by the means of the Gaussian components to produce intermediate designs.
- Outlier Detection: Identifying data points with low likelihood under the GMM distribution. These data points may represent novel or outlier designs that deviate significantly from the dataset's typical patterns.

In this project, the generation of novel design relied on the method of random sampling. By leveraging this approach, new and innovative designs were created by randomly drawing samples from the learned Gaussian Mixture Model (GMM), ensuring diversity and exploration within the design space. This process facilitated the exploration of unconventional design configurations, fostering creativity and enabling the discovery of unique solutions that may not have been evident through traditional design methods.

## 8. Resistance Calculation

The new generated novel hull form can be optimized to get a better design. For this project only resistance of the hull form was optimized and other form of optimizations can also be incorporated in future works. To calculate the resistance of the hull form, Michell Integral was chosen as the linear wave solver.

The Michell Integral, functioning as a linear wave solver, has become an invaluable asset in calculating hull form resistance, especially when compared to conventional methods such as Computational Fluid Dynamics (CFD) and Finite Element Method (FEM) analysis. Its adoption is primarily fueled by several notable advantages [9]. Firstly, the Michell Integral approach provides a streamlined mathematical framework for wave resistance analysis, offering computational efficiency in contrast to the intricate numerical simulations of CFD and FEM. This simplicity translates into reduced demands on computational resources and time, facilitating quicker analyses. Moreover, the method's alignment with linear wave theory, which assumes small-amplitude waves and linearized flow conditions, renders it particularly suitable for practical scenarios, notably in the initial design phases where swift assessments of hull performance are crucial [11].

Additionally, its computational efficiency enables seamless parametric studies, enabling rapid evaluation of various design parameters' impacts on wave resistance and efficient hull form optimization. Furthermore, the Michell Integral method's analytical nature fosters transparency in comprehending the underlying physical principles of wave resistance calculations, enhancing result interpretation, and offering insights into hull form hydrodynamics. Despite the high-fidelity results that CFD and FEM analyses may offer for complex flow phenomena, the Michell Integral method remains highly reliable for numerous practical cases, particularly when dealing with conventional hull forms and moderate wave conditions.

The Michell Integral is a method used to analyze the wave resistance of a ship's hull. It was developed by John Henry Michell, an Australian mathematician, in the late 19th century [11]. The principle behind the Michell Integral is rooted in potential flow theory, which models fluid flow around an object without considering viscosity.

The Michell Integral works by making several fundamental assumptions about the fluid flow around a ship's hull and then integrating the pressure distribution along the wetted surface of the hull to determine wave resistance. Initially, it assumes that the fluid flow can be approximated by potential flow theory, where the fluid is considered inviscid (no viscosity) and irrotational (no vortices). As a ship moves through water, it generates waves at its bow and stern due to pressure differences between the forward and aft ends, resulting in wave resistance opposing the ship's motion. The Michell Integral integrates the pressure distribution along the wetted surface, representing the portion of the hull in contact with water, to calculate wave-making resistance [10]. This mathematical formulation provides with a tool to estimate wave resistance during the ship design process.

For calculating resistance of the hull form using Michell Integral [18], a python code was written taking reference from the codes given by Noah J. Bagazinski and Faez Ahmed [1] in their technical paper.

For calculating the resistance, the value chosen for speed was 10 knots and the draft was given as 5 meters. The total resistance of the hull form was calculated as the sum of wave drag and skin friction drag: $R_t = R_w + R_f$

The wave drag ($R_w$) was calculated using the Michell Integral. The formula used is:

$$C_w = \frac{R_w}{\frac{1}{2}\rho U^2 LOA^2}$$

The skin friction drag ($R_f$) was calculated using the formula given by ITTC [12]:

$$C_f = \frac{0.075}{(Log_{10}(Re) - 2)^2}$$

The desired value for resistance was changed in the code to modify the hull form accordingly. To achieve this, the code was run in a loop till the desired value of resistance was gained.

The main aim of the Michell Integral code is to predict the wave drag from a ship progressing at a steady speed, neglecting the effects of viscosity. The first step is to generate the point grid and the inputs which are given for predicting the wave drag coefficient [18].

Point grid is generated for accurately representing the geometry of the hull form in the computational model. The Z and X vectors are used for defining the spatial coordinates of the grid points used in solver. The X vector represents the longitudinal positions along the hull, ranging from 0 (start) to the length overall (LOA) of the submerged hull. The Z vector represents the vertical positions below the waterline, ranging from the negative draft (deepest point) to 0 (water surface).

The Y grid represents the offsets of the ship hull at each grid point. These offsets define the shape of the hull in two dimensions, perpendicular to the longitudinal and vertical axes. By discretizing the hull shape into a grid of points, the solver can approximate the hull geometry and compute wave drag more efficiently. The Y grid is essential for applying Michell Thin Ship Theory, which relies on the hull shape to predict wave drag accurately.

The waterline length (WL) is a fundamental parameter for ship hydrodynamics, representing the length of the hull in contact with the water at a given draft. It is crucial for accurately estimating the flow patterns around the ship hull and predicting wave drag.

The inputs to the Michell Integral code are X, Y, Z vectors which is taken from the point grid. The code calculates the value of wave drag coefficients for combination of eight speeds and four drafts. Thus, an array of 32 wave drag coefficients is obtained. The speed and draft are taken as input and fed into the code, at which it is desired to calculate the wave drag resistance. Another function is defined that takes in these speeds and drafts to interpolate between these to estimate the wave drag coefficient.

The 'CalcDrag' function takes various parameters related to the ship's geometry, speed, and environmental conditions to estimate the total drag force. The function returns a tuple containing the calculated total drag force and the Froude number (Fn). The total drag force includes contributions from both wave drag and skin friction drag.

The function begins by calculating the skin friction coefficient (Cf) using the Calc_Cf function, which estimates the viscous drag based on the ship's speed and waterline length. Next, it computes the skin friction drag (Rf) using the formula for drag force, considering the skin friction coefficient, water density, wetted surface area, and ship speed. It calculates the Froude number (Fn), which is a dimensionless parameter representing the ratio of inertial forces to gravitational forces and is crucial for hydrodynamic analysis. The function then interpolates the wave drag coefficient (Cw) based on the Froude number (Fn) and draft fraction (T) using the interp_CW function. Using the interpolated wave drag coefficient, it computes the wave drag force (Rw) considering water density, ship length overall squared, and ship speed squared. Finally, the function returns the sum of wave drag and skin friction drag as the total drag force experienced by the ship.

## 9. Generating New Dataset

Multiple random samples were generated after applying GMM to the training dataset. Resistance of each these hull forms were calculated using code written in Michell Integral. The total resistance was calculated as the summation of the wave drag resistance and skin friction resistance. The wave drag resistance was calculated using the Michell Integral and skin friction resistance was calculated using the formula given by ITTC. For calculating the wave drag resistance, the speed of the hull form moving through the inviscid fluid was taken as 10 knots and the draft was taken as 5 m.

After calculating the total resistance of each hull form of the generated dataset, hull forms with resistance less than 5000 N were grouped and saved in a different CSV file. A total of 150 hull forms were generated. The geometry of each hull form in this CSV file is represented by 45 parameters. The advantage of creating such a generative model is that a new dataset of hull forms can be generated as per the constraints given by the user, and then accordingly select the most feasible hull form as per the given requirements.

For visualizing this generated dataset, again t-SNE was used. From the figure given below, it was observed that the generated dataset is within the distribution of the training dataset.

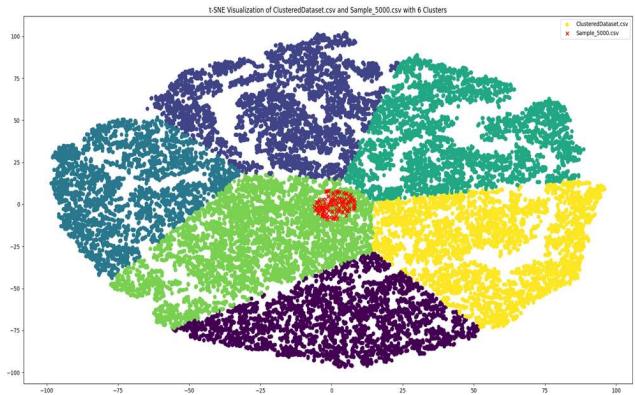

Figure 13: New Dataset t-SNE

Some of the samples which were generated are shown below.

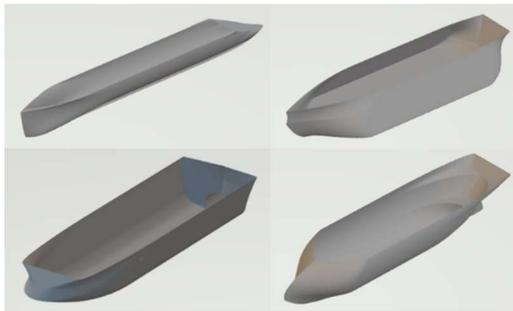

Figure 14: Novel Generated Designs

Inputs Given to obtain this dataset:

- Resistance < 5000 N
- Speed = 10 knots
- Density of water = 1025 kg/m$^3$
- Draft = 0.6 * Depth of the hull

## Conclusion

This project report outlines a comprehensive approach to designing and optimizing ship hull forms using generative AI techniques and traditional hydrodynamic analysis methods. The conventional approach to ship design involves a time-consuming process that heavily relies on human expertise and iterative refinement. However, the emergence of generative AI offers a paradigm shift by leveraging computational algorithms to explore a broader design space, generate novel hull forms, and automate the optimization process.

By utilizing a dataset comprising 30,000 hull forms and training a Gaussian Mixture Model (GMM) on this data, the generative AI can learn underlying patterns and relationships within the dataset, enabling the generation of novel and unconventional hull designs. This approach significantly reduces the time and effort required for design optimization compared to traditional methods, offering a multidisciplinary optimization framework that integrates various design objectives and constraints across disciplines like hydrodynamics, structures, and stability.

Moreover, the report highlights the use of the Michell Integral as a linear wave solver to calculate hull resistance, emphasizing its advantages over conventional methods like Computational Fluid Dynamics (CFD) and Finite Element Method (FEM) analysis. The

Michell Integral provides a streamlined mathematical framework for wave resistance analysis, offering computational efficiency, applicability to linear wave theory, suitability for parametric studies, transparency, and reliability for practical cases.

Overall, the integration of generative AI techniques and traditional hydrodynamic analysis methods presents a promising approach to ship hull design and optimization. By harnessing the power of computational algorithms and machine learning, naval architects and marine engineers can accelerate the design process, explore innovative design solutions, and enhance the efficiency and performance of ship hulls. This project report lays the foundation for future advancements in ship design methodologies, paving the way for more sustainable, efficient, and technologically advanced vessels.

## References

[1] Noah J. Bagazinski and Faez Ahmed, 2023. "SHIP-D: Ship Hull Dataset for Design Optimization Using Machine Learning." The MIT Press.

[2] Young-Soon Yang, Chang-Kue Park, Kyung-Ho Lee and Jung-Chun Suh, 2007. "A study on the preliminary ship design method using deterministic approach and probabilistic approach including hull form." Springer-Verlag.

[3] Evans, J. H., 1959. "Basic design concepts." Journal of the American Society for Naval Engineers.

[4] Digital Blue Foam. "Generative Architecture: What Does It Entail?" Blog DBF. Available at: https://www.digitalbluefoam.com/post/generative-architecture-what-does-it-entail

[5] MDPI. "Simplified Methods for Generative Design That Combine Evaluation Techniques for Automated Conceptual Building Design." Available at: https://www.mdpi.com/2076-3417/13/23/12856

[6] YouTube. "What is Generative Design in Architecture?" Available at: https://www.youtube.com/watch?v=8KQ0osjxXIg

[7] Oscar Contreras Carrasco. "Gaussian Mixture Models Explained." Towards Data Science. Available at: https://towardsdatascience.com/gaussian-mixture-models-explained-6986aaf5a95

[8] Sculpteo. "What is an STL file?" Sculpteo 3D Learning Hub. Available at: https://www.sculpteo.com/en/3d-learning-hub/create-3d-file/what-is-an-stl-file/

[9] A. Sh. Gotman, 2002. "Study Of Michell's Integral And Influence Of Viscosity And Ship Hull Form On Wave Resistance." Oceanic Engineering International, Vol. 6, No. 2, pp. 74-115. Novosibirsk State Academy of Water Transport, Shchetinkina st., 33, Novosibirsk, 630099, Russia.

[10] J. H. Michell, 1898. "The Wave-Resistance of a Ship." Philosophical Magazine, vol. 45, Ser. 5, pp. 106–123. John Henry Michell (1863–1940), Australian mathematician and Professor of Mathematics at the University of Melbourne.

[11] E. O. Tuck, 1989. "The Wave Resistance Formula of J.H. Michell (1898) and Its Significance to Recent Research in Ship Hydrodynamics." J. Austral. Math. Soc.

Ser. B 30, pp. 365-377. Received 28 July 1988; revised 26 August 1988.

[12] ITTC, 2002. "ITTC – Recommended Procedures 7.5-02 - Resistance Uncertainty Analysis, Example for Resistance Test." Effective Date 2002, Revision 01, Page 11 of 17.

[13] Deepchecks Community Blog. (September 11, 2023). "How to Train Generative AI Models." Deepchecks Community Blog.
Available at: https://deepchecks.com/how-to-train-generative-ai-models/

[14] TechTarget. (n.d.). Generative AI: What is generative AI? Everything you need to know. (2024, February 2).
Available at: https://www.techtarget.com/searchenterpriseai/definition/generative-AI

[15] Towards Data Science. (2024, February 2). Clustering Based Unsupervised Learning.
Available at: https://towardsdatascience.com/clustering-based-unsupervised-learning-8d705298ae51

[16] Hex. (2024, January 23). Unveiling patterns in unlabeled data with k-means clustering.
Available at: https://hex.tech/blog/Unveiling-patterns-in-unlabeled-data-with-k-means-clustering/

[17] Enjoy Algorithms. t-SNE (t-Distributed Stochastic Neighbor Embedding) Algorithm. (2024, March 3).
Available at: https://www.enjoyalgorithms.com/blog/tsne-algorithm-in-ml

[18] Douglas Read, 2009. "A Drag Estimate for Concept-Stage Ship Design Optimization." DigitalCommons@UMaine, Electronic Theses and Dissertations, Fogler Library.

[19] Khalid, Z. (2022, January 17). Fast-Track Construction of Indian Navy's Stealth Destroyers. Centre for Strategic and Contemporary Research. Retrieved from https://cscr.pk/explore/themes/defense-security/fast-track-construction-of-indian-navys-stealth-destroyers/